\documentclass{article}
\usepackage{spconf,amsmath,graphicx}
\usepackage{amsfonts}
\usepackage{amsmath}
\usepackage{adjustbox}
\usepackage{algorithmicx}
\usepackage{algpseudocode}
\usepackage[ruled]{algorithm}
\usepackage{graphicx}
\usepackage{caption}
\usepackage{mathtools}

\usepackage{graphicx,times,amsmath} % Add all your packages here
\usepackage{caption}
\usepackage{subcaption}
\usepackage[rightcaption]{sidecap}
\usepackage{caption}
\usepackage{multirow}
\usepackage{hhline}
\usepackage{amsmath}
\usepackage{epstopdf}
\usepackage{tabularx}
\usepackage{makecell}

\graphicspath{ {images/} }

\usepackage{alphalph}

\title{Image Augmentation using Radial Transform for \\Training Deep Neural Networks}
\name{Hojjat~Salehinejad$^{\star \dagger}$,~Shahrokh~Valaee$^{\star}$, Tim~Dowdell$^{\dagger}$,~and~Joseph~Barfett$^{\dagger}$\thanks{The authors thank the support of NVIDIA Corporation with the
donation of the Titan X GPUs used for this project.}}
\address{$^{\star}$Department of Electrical \& Computer Engineering, University of Toronto, Toronto, Canada \\
$^{\dagger}$Department of Medical Imaging, St. Michael's Hospital, University of Toronto, Toronto, Canada \\
\textit{salehinejadh@smh.ca, valaee@ece.utoronto.ca, \{dowdellt,barfettj\}@smh.ca}}

\usepackage{fancyhdr}
\fancypagestyle{pageStyleOne}{%
\normalfont\footnotesize
    \fancyhf{}
\fancyhead[L]{\fontsize{8}{8} \selectfont This paper is accepted for presentation at IEEE International Conference on Acoustics, Speech and Signal Processing (IEEE ICASSP), 2018.}
}

\begin{document}
\newcommand*{\img}{%
  \includegraphics[
    width=\linewidth,
    height=20pt,
    keepaspectratio=false,
  ]{example-image-a}%
}

\maketitle
\thispagestyle{pageStyleOne}
\begin{abstract}
Deep learning models have a large number of free parameters that must be estimated by efficient training of the models on a large number of training data samples to increase their generalization performance. In real-world applications, the data available to train these networks is often limited or imbalanced. We propose a sampling method based on the radial transform in a polar coordinate system for image augmentation to facilitate the training of deep learning models from limited source data. This pixel-wise transform provides representations of the original image in the polar coordinate system by generating a new image from each pixel. This technique can generate radial transformed images up to the number of pixels in the original image to increase the diversity of poorly represented image classes. Our experiments show improved generalization performance in training deep convolutional neural networks with radial transformed images.
\end{abstract}

%Deep learning models are complex architectures composed of many layers of artificial neurons, resulting in a large number of free parameters that must be estimated by training the network. These models typically require a large number of training data samples for high generalization performance. In real-world applications, the data available to train these networks is often limited or imbalanced. We propose a sampling method based on radial transform in polar coordinate space for image augmentation to facilitate the training of neural networks from limited source data. This pixel-wise transform provides representations of the original image in the polar coordinate system by generating a new image from each pixel of the original image. This technique can generate radial transformed images up to the number of pixels in the original image to increase the diversity of poorly represented image classes. Experiments performed on MNIST and a set of naturally imbalanced medical images show improved generalization performance in training deep convolutional neural networks with radial transformed images comparing with training the same networks but with original little images or Affine transformed images.

% Note that keywords are not normally used for peerreview papers.

\section{Introduction}
\label{sec:intro}
%Deep learning refers to a network of connected artificial neurons in multiple layers, which can perform feature extraction from observed data and learn the complicated relationships among the features of data. These models generally use non-linear
%but simple units, which can provide different levels of feature
%representation through their multi-layer architecture. The higher
%layers provide a more abstract representation of data and
%suppress irrelevant variations \cite{lecun2015deep}. 

The need for massive amounts of data to train deep neural networks is a major drawback to these models~\cite{zhou2006training}, \cite{salehinejad2017recent}. Generalization performance and versatility of deep learning \cite{lecun2015deep} models are highly dependent on availability of abundant data. The generalization performance refers
to the accuracy of the neural network in classification of unseen data. The other challenge is imbalanced datasets, where very few data samples are available for some data classes~\cite{salehinejad2018generalization}. These challenges arise in many practical machine learning scenarios such as financial transactions~\cite{salehinejad2016customer} and fraud detection in banking transactions or in medical sciences~\cite{salehinejad2018generalization},~\cite{pouladi2015recurrent}. In the former, a small number of fraudulent transactions are imbalanced by a high percentage of normal transactions. In the latter, the majority of the population being healthy or a low prevalence rate for certain medical conditions in the dataset can bias the deep learning model. 

The training of a neural network with limited data may be mitigated by sampling noise, which exists in the training data but not in the test data drawn from the same distribution \cite{srivastava2014dropout}. An elegant solution to these challenges is data augmentation, i.e., the application of one or more deformations to a collection of annotated training samples which result in new, additional, and potentially non-redundant training data \cite{krizhevsky2012imagenet}, \cite{salamon2017deep}. 

In general, data augmentation does not increase the information content of the dataset. However, it can improve diversity of the dataset and generalization performance. Diversifying the data helps the network to generalize better to unseen data and become invariant to applied deformations \cite{salamon2017deep}. The neural network learns from the added diversity and gains experience in how data belonging to the given labels can ``look different". The various deformations commonly applied to labeled data, such as multiplication by a transform matrix, does not affect the semantic meaning of the labels \cite{salamon2017deep}. Some of the image augmentation techniques include adding noise, rotating, translating, mirroring, or scaling the image. Affine is a 2D geometric transform method based on applying a combination of translation, rotation, scaling, and shearing transformations. The Affine augmentation method is widely used as an image augmentation method for correcting geometric distortion introduced by perspective irregularities \cite{stearns1995method}. Another approach is making an image that contains multiple copies of the original image rotated by different angles \cite{okaforoperational}. The polar harmonic transform, based on a set of orthogonal projections, is another method that has been used to generate a set of features that are insensitive to rotation \cite{yap2010two}.

%Spherical images augmented with depth information and a spherical saliency map were developed for vision-based navigation in real urban environments \cite{meilland2010spherical}. The polar harmonic transform, based on a set of orthogonal projections, has been used to generate a set of features that are insensitive to rotation \cite{yap2010two}. The log-polar transform has been used for image registration \cite{matungka2009image}, \cite{zhang2016forward}, whereby the transform can eliminate the rotation and scale effects on an image by constructing a new representation in the polar coordinate system. Hierarchical kernels then can eliminate the row and column translation effects from the new image \cite{tang2014hierarchical}. 

\begin{figure}[th]
\captionsetup{font=small}

\centering
        \begin{subfigure}[t]{0.23\textwidth}
        \centering
\includegraphics[width=1.\textwidth]{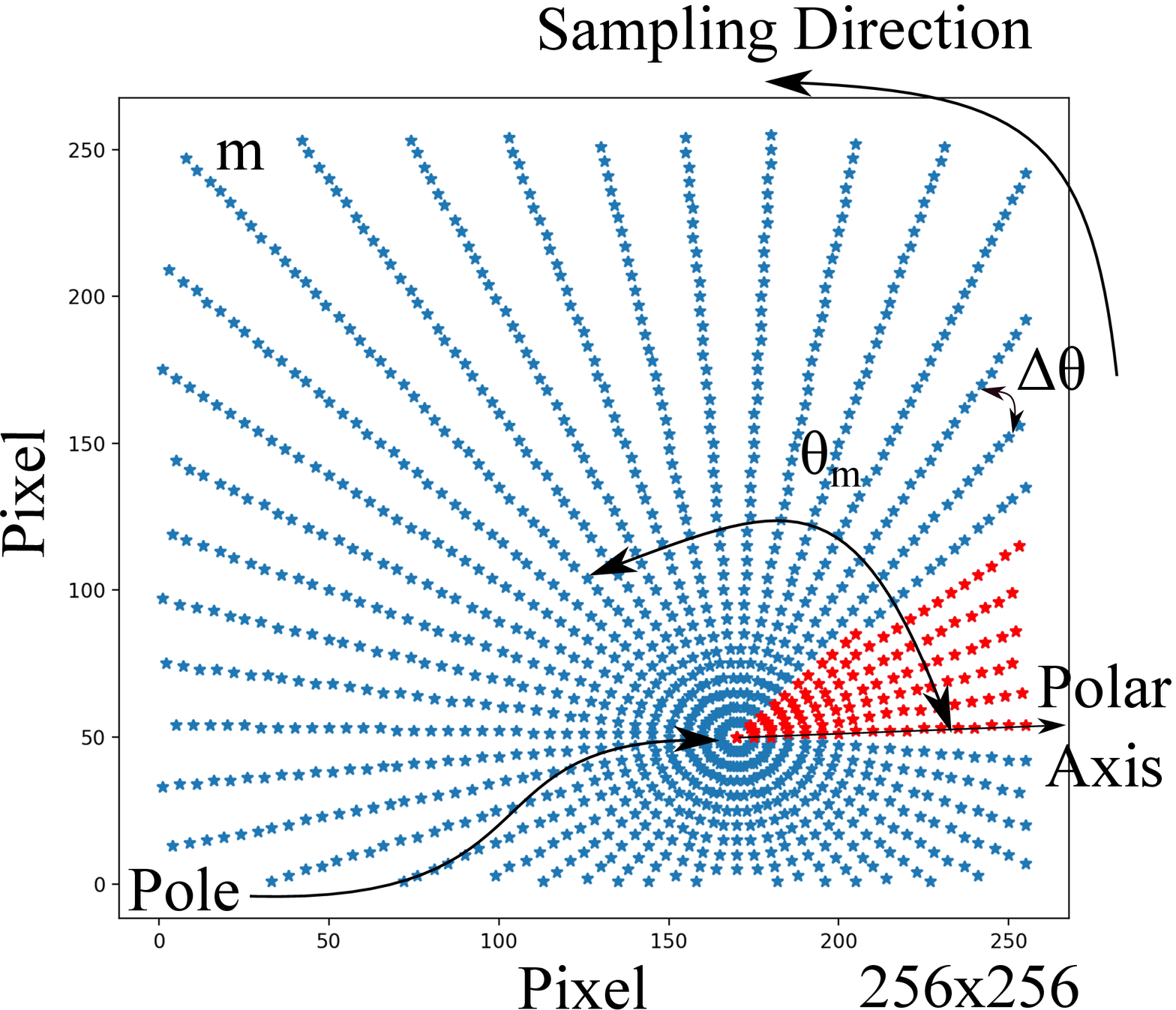}
                \caption{Sampling from pixels using radial transform.}
                \label{fig:samplingb}
        \end{subfigure}%   
        ~~
\begin{subfigure}[t]{0.23\textwidth}
        \centering
\includegraphics[width=1.\textwidth]{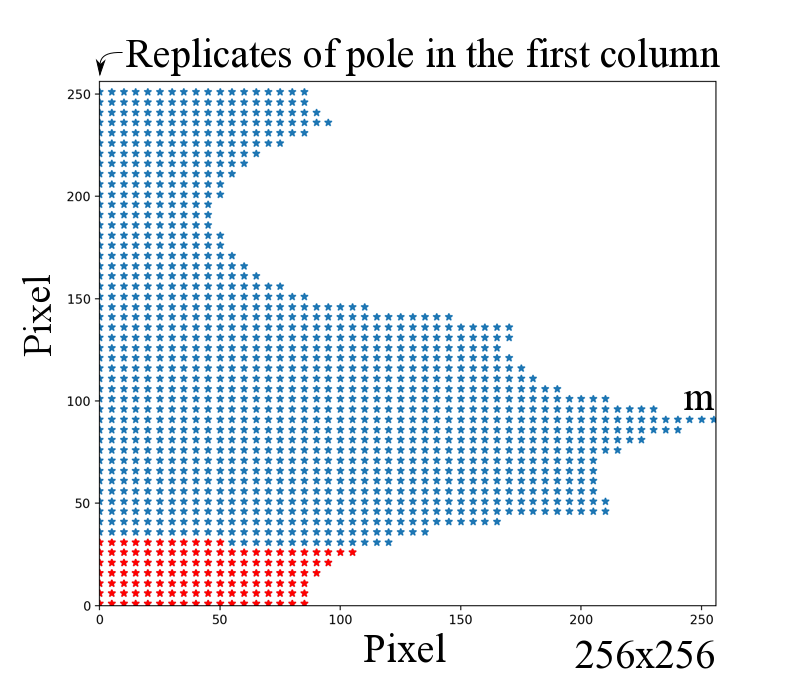}
                \caption{Constructed image from selected samples in the polar coordinate system.}
                \label{fig:samplingc}
        \end{subfigure}%   
        
        \begin{subfigure}[t]{0.23\textwidth}
        \centering
\includegraphics[width=0.78\textwidth]{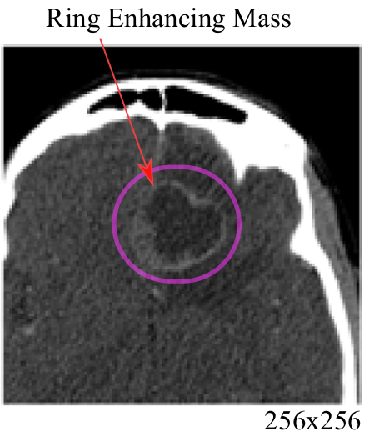}
                \caption{Ring enhancing mass on a brain MRI.}
                \label{fig:splitpage}
        \end{subfigure}%   
        ~~
\begin{subfigure}[t]{0.23\textwidth}
        \centering
\includegraphics[width=0.76\textwidth]{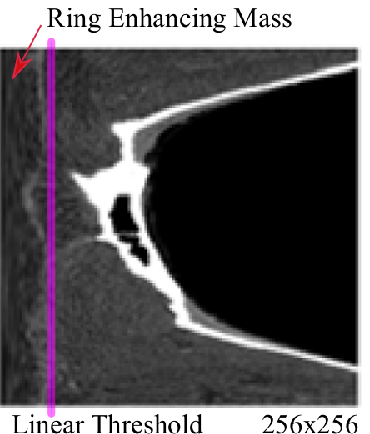}
                \caption{Augmented MRI from original MRI using radial transform.}
                \label{fig:samplinga}
        \end{subfigure}%  
        \vspace{-2mm}
\caption{Radial transform sampling. a) Selected discrete samples using radial transform on an arbitrary plane. The arbitrary selected pole is at pixel $O(170,50)$; b) Mapping of selected samples in (a) from polar coordinate system to Cartesian coordinate system. The red samples show the direction of mapping the samples in (a) into (b); c) An example of radial transform application in tumor detection: Ring enhancing mass on an original brain MRI. Detection of the mass requires non-linear threshold(s); d) The original brain MRI in (c) after radial transform with pole selected in the middle of the mass. The ring enhancing mass is expanded (up-sampled) and can be detected with a linear threshold. }
\label{fig:sampling}
\vspace{-4mm}
\end{figure}
\alglanguage{pseudocode}

In this paper, an augmentation method based on the radial transform is proposed to sample images in the polar coordinate system and map the samples to Cartesian space for construction of new (augmented) images. A radial transformed image is a coherent representation of the original image and maintains the semantic validity of the data classes.

\section{Proposed Method}
\label{sec:methods}

We define a point on a plane in the polar coordinate system as $(r,\theta)\in \mathcal{P}^{2}$ where $r\in \mathbb{Z}^{+}$ is the radial coordinate and $\theta\in \mathbb{R}^{+}$ is the counterclockwise angular coordinate with respect to a polar axis drawn horizontally from the pole to the right, as illustrated in Figure~\ref{fig:sampling}(a). We draw $M$ distinct rays with length $R_{m}\in \mathbb{R}^{+}$ where $m=0$ is the arbitrary initial ray along the polar axis. Each ray has an identical angular difference $\Delta\theta=|\theta_{m+1}-\theta_{m}|$ from its adjacent ray. Then, we generate a set of spatial coordinates $K=\{(r,\theta_{m})_{k}\}$ in $\mathcal{P}^{2}$, with respect to a pole $O(u,v)\in\mathcal{P}^{2}$ such that $u,v\in \mathbb{Z}^{+}$, which is an arbitrary selected pixel in the original image $\textbf{X}_{M\times N}\in \mathcal{C}^{2}$ in the Cartesian coordinate system.
Therefore, to generate a point $(r,\theta_{m})_{k}$ for $r\in \{0,...,N-1\}$ we have
\begin{equation}
\theta_{m} = 2\:\pi \cdot m/M,
\end{equation}
where $m\in\{0,...,M-1\}$.
By $\textbf{X} \xrightarrow{\phi (r,\theta_{m})_{k}} \hat{\textbf{X}}$, we map the pixels at Cartesian coordinates $(m,n)\in \mathbb{Z}^{+}$ from the original image $\textbf{X}$ in $\mathcal{C}^{2}$ to construct the augmented image $\hat{\textbf{X}}\in \mathcal{C}^{2}$ in the Cartesian space using radial transform $\phi(\cdot)$ as
\begin{equation}
\hat{x} = round(r \cdot cos(\theta_{m})) \:\:  \& \:\:  \hat{y} = round(r \cdot sin(\theta_{m}))
\end{equation}
for $r\in \{0,...,N-1\}$ and $m\in \{0,...,M-1\}$ such that~${0\leq u+\hat{x} < M}$ and ${0 \leq  v+\hat{y} < N}$. These conditions guarantee that the pair $(\hat{x},\hat{y})$ stays spatially within $\hat{\textbf{X}}$. 
A new pixel $(m,r)$ in the constructed image is then defined as ${\hat{x}_{m,r}=x_{u+\hat{x}, v+\hat{y}}}.$
The image $\tilde{\textbf{X}}$ is the radial transform of $\textbf{X}$ with respect to the pixel $O(u,v)\in \textbf{X}$. 

The pole at pixel location $O(u, v)$ in $\textbf{X}$ is repeated as the first pixel in every row $m\in\{0,...,M-1\}$ of $\hat{\textbf{X}}$, as illustrated in Figure~\ref{fig:sampling}(b). As $r\to N$ for an arbitrary $\theta_{m}$, pixels in close neighborhood of $O(u,v)$ are up-sampled and pixels in further neighborhood are down-sampled. The proposed radial transform can generate unique radial transformed images up to the number of pixels in $\textbf{X}$, which is $M\times N$. Such high diversity of images preserves the dependencies among local and global pixels in $\textbf{X}$ but in different representations. For a typical $256\times 256$ original image, the rotation augmentation approach generates much less new representations of the original image (e.g., 360 new representations by having a rotation step of one degree) comparing with the radial transform, which generates $256\times 256=65,536$ new images.

Figure~\ref{fig:sampling}(c) shows a practical example on how radial transform can help to map a set of pixels belonging to a ring enhancing mass in a brain magnetic resonance imaging (MRI) to be mapped into a new image using polar coordinate system. The pole is selected in the middle of the mass. Distinguishing the mass in the Cartesian plane requires a set of non-linear thresholds. However, radial transform in Figure~\ref{fig:sampling}(d) has expanded (i.e., up-sampled) the image in local neighborhood of the mass and has compressed (down-sampled) the further, less important pixels. In this way the mass can be easily separated with a linear threshold. This example clearly shows how radial transform not only can help to augment images but also can reduce complexity of classification problems in images. In this paper, our focus is on the augmentation property of the radial transform.

\begin{figure*}[t]
\centering
\captionsetup{font=small}
        \begin{subfigure}[t]{0.07\textwidth}
        \centering
                \includegraphics[width=0.95\textwidth]{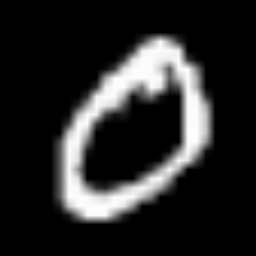}
                \caption{}
                \label{fig:loss}
        \end{subfigure}%   
        %\hspace{-1mm}
        \begin{subfigure}[t]{0.07\textwidth}
        \centering
                \includegraphics[width=0.95\textwidth]{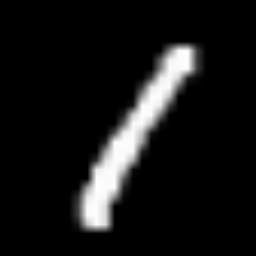}
                \caption{}
                \label{fig:loss}
        \end{subfigure}%   
         \begin{subfigure}[t]{0.07\textwidth}
        \centering
                \includegraphics[width=0.95\textwidth]{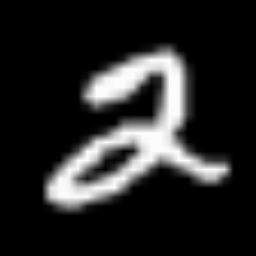}
                \caption{}
                \label{fig:loss}
        \end{subfigure}%   
         \begin{subfigure}[t]{0.07\textwidth}
        \centering
                \includegraphics[width=0.95\textwidth]{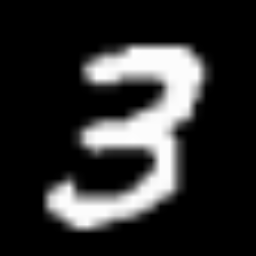}
                \caption{}
                \label{fig:loss}
        \end{subfigure}%   
        \begin{subfigure}[t]{0.07\textwidth}
        \centering
                \includegraphics[width=0.95\textwidth]{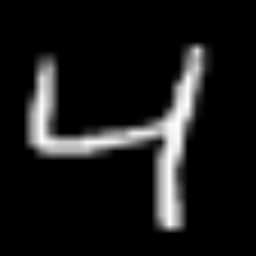}
                \caption{}
                \label{fig:loss}
        \end{subfigure}%   
        \begin{subfigure}[t]{0.07\textwidth}
        \centering
                \includegraphics[width=0.95\textwidth]{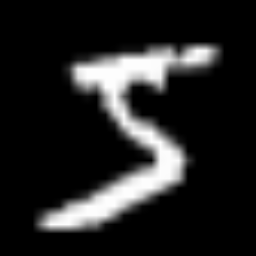}
                \caption{}
                \label{fig:loss}
        \end{subfigure}%   
        \begin{subfigure}[t]{0.07\textwidth}
        \centering
                \includegraphics[width=0.95\textwidth]{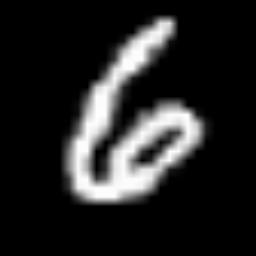}
                \caption{}
                \label{fig:loss}
        \end{subfigure}% 
        \begin{subfigure}[t]{0.07\textwidth}
        \centering
                \includegraphics[width=0.95\textwidth]{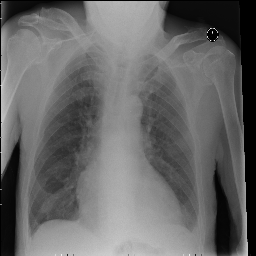}
                \caption{}
                \label{fig:loss}
        \end{subfigure}%   
         \begin{subfigure}[t]{0.07\textwidth}
        \centering
                \includegraphics[width=0.95\textwidth]{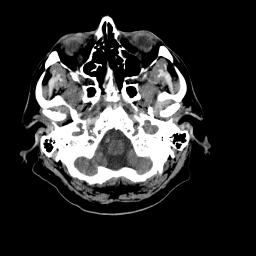}
                \caption{}
                \label{fig:loss}
        \end{subfigure}%   
           \begin{subfigure}[t]{0.07\textwidth}
        \centering
                \includegraphics[width=0.95\textwidth]{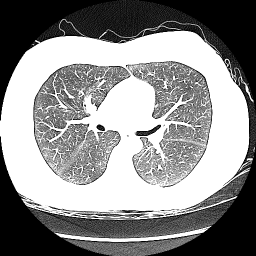}
                \caption{}
                \label{fig:loss}
        \end{subfigure}%   
        \begin{subfigure}[t]{0.07\textwidth}
        \centering
                \includegraphics[width=0.95\textwidth]{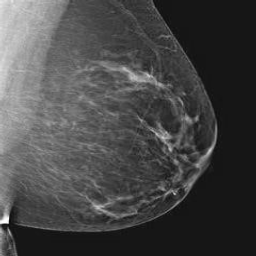}
                \caption{}
                \label{fig:loss}
        \end{subfigure}%   
         \begin{subfigure}[t]{0.07\textwidth}
        \centering
                \includegraphics[width=0.95\textwidth]{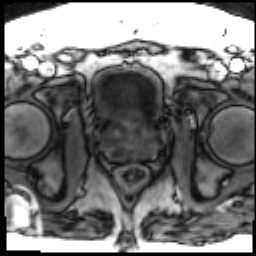}
                \caption{}
                \label{fig:loss}
        \end{subfigure}%      
                 \begin{subfigure}[t]{0.07\textwidth}
        \centering
                \includegraphics[width=0.95\textwidth]{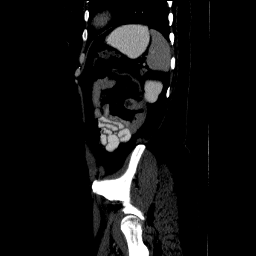}
                \caption{}
                \label{fig:loss}
        \end{subfigure}%   
         \begin{subfigure}[t]{0.07\textwidth}
        \centering
                \includegraphics[width=0.95\textwidth]{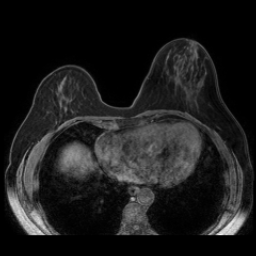}
                \caption{}
                \label{fig:loss}
        \end{subfigure}%   

           \begin{subfigure}[t]{0.07\textwidth}
        \centering
                \includegraphics[width=0.95\textwidth]{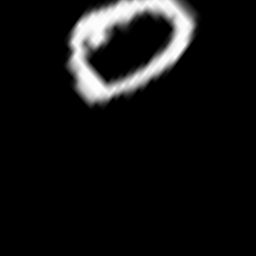}
                \caption{$\psi$(a)}
                \label{fig:loss}
        \end{subfigure}%   
        \begin{subfigure}[t]{0.07\textwidth}
        \centering
                \includegraphics[width=0.95\textwidth]{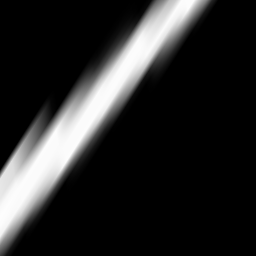}
                \caption{$\psi$(b)}
                \label{fig:loss}
        \end{subfigure}%   
         \begin{subfigure}[t]{0.07\textwidth}
        \centering
                \includegraphics[width=0.95\textwidth]{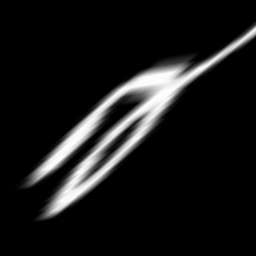}
                \caption{$\psi$(c)}
                \label{fig:loss}
        \end{subfigure}%   
         \begin{subfigure}[t]{0.07\textwidth}
        \centering
                \includegraphics[width=0.95\textwidth]{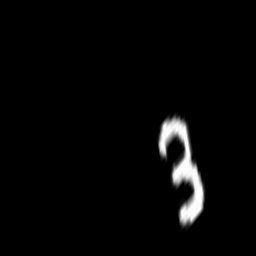}
                \caption{$\psi$(d)}
                \label{fig:loss}
        \end{subfigure}%   
        \begin{subfigure}[t]{0.07\textwidth}
        \centering
                \includegraphics[width=0.95\textwidth]{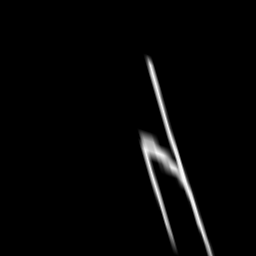}
                \caption{$\psi$(e)}
                \label{fig:loss}
        \end{subfigure}%   
        \begin{subfigure}[t]{0.07\textwidth}
        \centering
                \includegraphics[width=0.95\textwidth]{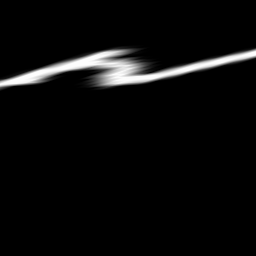}
                \caption{$\psi$(f)}
                \label{fig:loss}
        \end{subfigure}%   
        \begin{subfigure}[t]{0.07\textwidth}
        \centering
                \includegraphics[width=0.95\textwidth]{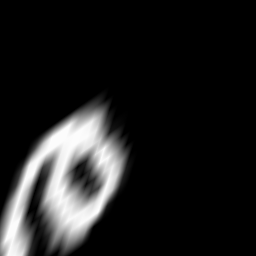}
                \caption{$\psi$(g)}
                \label{fig:loss}
        \end{subfigure}%   
        \begin{subfigure}[t]{0.07\textwidth}
        \centering
                \includegraphics[width=0.95\textwidth]{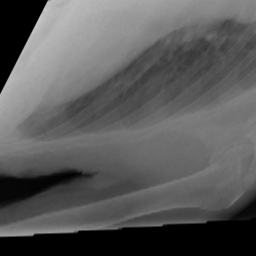}
                \caption{$\psi$(h)}
                \label{fig:loss}
        \end{subfigure}%   
         \begin{subfigure}[t]{0.07\textwidth}
        \centering
                \includegraphics[width=0.95\textwidth]{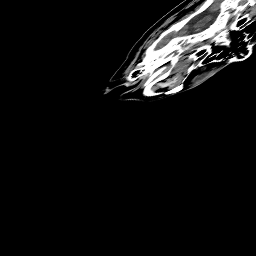}
                \caption{$\psi$(i)}
                \label{fig:loss}
        \end{subfigure}%   
           \begin{subfigure}[t]{0.07\textwidth}
        \centering
                \includegraphics[width=0.95\textwidth]{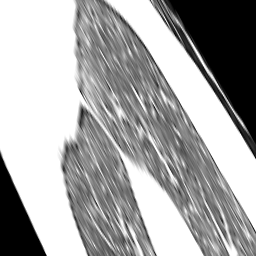}
                \caption{$\psi$(j)}
                \label{fig:loss}
        \end{subfigure}%   
        \begin{subfigure}[t]{0.07\textwidth}
        \centering
                \includegraphics[width=0.95\textwidth]{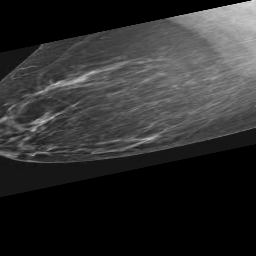}
                \caption{$\psi$(k)}
                \label{fig:loss}
        \end{subfigure}%   
         \begin{subfigure}[t]{0.07\textwidth}
        \centering
                \includegraphics[width=0.95\textwidth]{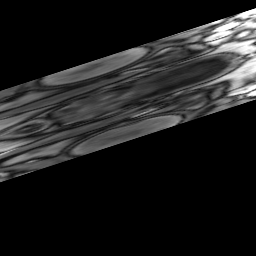}
                \caption{$\psi$(l)}
                \label{fig:loss}
        \end{subfigure}%   
         \begin{subfigure}[t]{0.07\textwidth}
        \centering
                \includegraphics[width=0.95\textwidth]{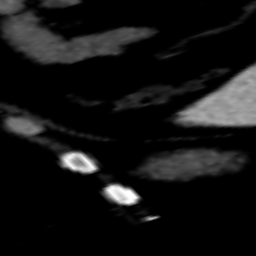}
                \caption{$\psi$(m)}
                \label{fig:loss}
        \end{subfigure}%   
         \begin{subfigure}[t]{0.07\textwidth}
        \centering
                \includegraphics[width=0.95\textwidth]{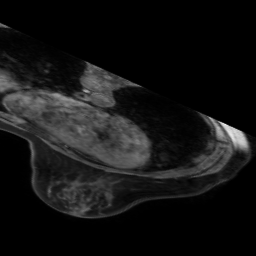}
                \caption{$\psi$(n)}
                \label{fig:loss}
        \end{subfigure}%  
             
           \begin{subfigure}[t]{0.07\textwidth}
        \centering
                \includegraphics[width=0.95\textwidth]{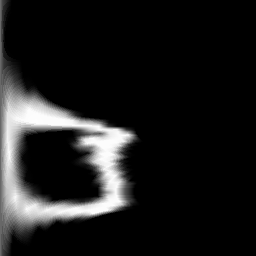}
                \caption{$\phi$(a)}
                \label{fig:loss}
        \end{subfigure}%   
        \begin{subfigure}[t]{0.07\textwidth}
        \centering
                \includegraphics[width=0.95\textwidth]{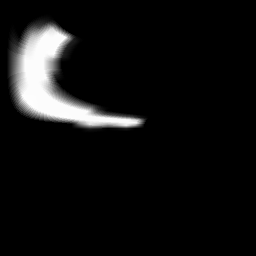}
                \caption{$\phi$(b)}
                \label{fig:loss}
        \end{subfigure}%   
         \begin{subfigure}[t]{0.07\textwidth}
        \centering
                \includegraphics[width=0.95\textwidth]{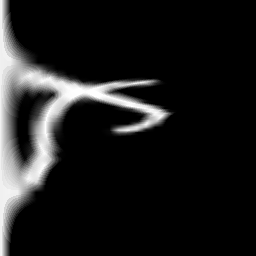}
                \caption{$\phi$(c)}
                \label{fig:loss}
        \end{subfigure}%   
         \begin{subfigure}[t]{0.07\textwidth}
        \centering
                \includegraphics[width=0.95\textwidth]{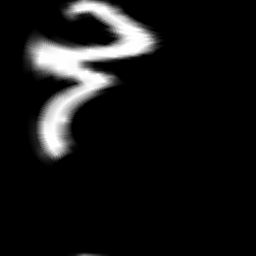}
                \caption{$\phi$(d)}
                \label{fig:loss}
        \end{subfigure}%   
        \begin{subfigure}[t]{0.07\textwidth}
        \centering
                \includegraphics[width=0.95\textwidth]{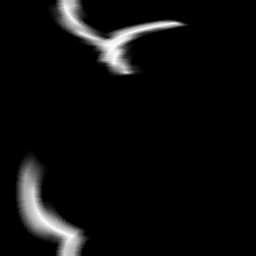}
                \caption{$\phi$(e)}
                \label{fig:loss}
        \end{subfigure}%   
        \begin{subfigure}[t]{0.07\textwidth}
        \centering
                \includegraphics[width=0.95\textwidth]{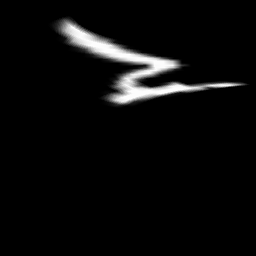}
                \caption{$\phi$(f)}
                \label{fig:loss}
        \end{subfigure}% 
        \begin{subfigure}[t]{0.07\textwidth}
        \centering
                \includegraphics[width=0.95\textwidth]{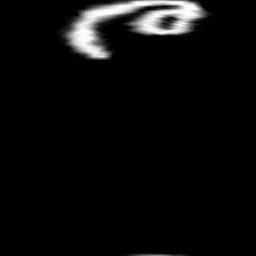}
                \caption{$\phi$(g)}
                \label{fig:loss}
        \end{subfigure}% 
        \begin{subfigure}[t]{0.07\textwidth}
        \centering
                \includegraphics[width=0.95\textwidth]{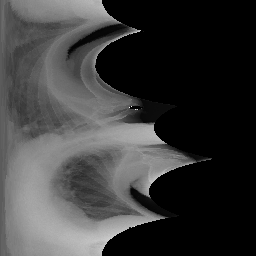}
                \caption{$\phi$(h)}
                \label{}
        \end{subfigure}%   
         \begin{subfigure}[t]{0.07\textwidth}
        \centering
                \includegraphics[width=0.95\textwidth]{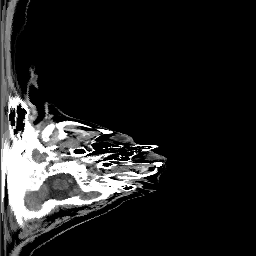}
                \caption{$\phi$(i)}
                \label{fig:loss}
        \end{subfigure}%   
           \begin{subfigure}[t]{0.07\textwidth}
        \centering
                \includegraphics[width=0.95\textwidth]{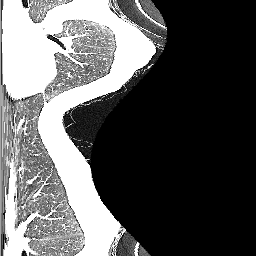}
                \caption{$\phi$(j)}
                \label{fig:loss}
        \end{subfigure}%   
        \begin{subfigure}[t]{0.07\textwidth}
        \centering
                \includegraphics[width=0.95\textwidth]{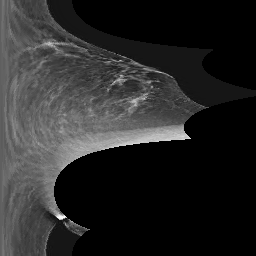}
                \caption{$\phi$(k)}
                \label{fig:loss}
        \end{subfigure}%   
         \begin{subfigure}[t]{0.07\textwidth}
        \centering
                \includegraphics[width=0.95\textwidth]{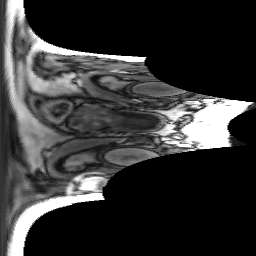}
                \caption{$\phi$(l)}
                \label{fig:loss}
        \end{subfigure}%   
                 \begin{subfigure}[t]{0.07\textwidth}
        \centering
                \includegraphics[width=0.95\textwidth]{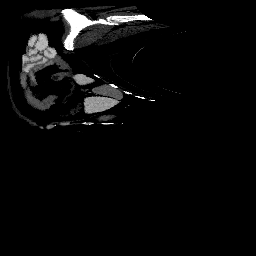}
                \caption{$\phi$(m)}
                \label{fig:loss}
        \end{subfigure}%   
         \begin{subfigure}[t]{0.07\textwidth}
        \centering
                \includegraphics[width=0.95\textwidth]{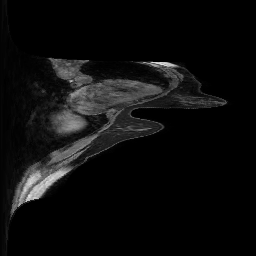}
                \caption{$\phi$(n)}
                \label{fig:loss}
        \end{subfigure}%        
        \vspace{-2mm}
        \caption{A sample from some data classes of MNIST and medical multimodal datasets with their corresponding representation using radial transform, ($\phi(\cdot)$) and Affine transform ($\psi(\cdot)$), for a randomly selected pixel. a) 0; b) 1; c) 2; d) 3; e) 4; f) 5; g) 6; h) Chest X-Ray; i) Head CT; j)~Lung CT; k)~Mammogram; l) Pelvis MRI; m) Sagittal Abdomen MRI; n) Breast MRI.}
        \label{fig:mnist} 
                \vspace{-3mm}
\end{figure*}

A neural network can be trained with the generated radial transformed images. An image can be classified by applying $\phi(\cdot)$ on a subset of pixels $T$ and detect the corresponding classification labels $L=\{c_{t}: c_{t}=arg\:max\{p_{t,1},...,p_{t,C}\} \; \forall \; t \in T\}$ using a trained network with radial transformed images. These labels can be directly used for segmentation and multi-object detection applications. As an example for the single-object image classification task, the majority of votes from the predicted labels (i.e., the most frequent label in $L$) is the predicted class

\section{Experiments}
\label{sec:results}
Experiments were conducted using GoogLeNet  \cite{szegedy2015going} and AlexNet \cite{krizhevsky2012imagenet} on two datasets: the MNIST dataset with 10 classes of hand written digits \cite{lecun2010mnist} and a dataset of naturally limited 9 different modalities of medical images \cite{cancerwebsite}. A sample from some of the image classes is presented in the first row of Figure~\ref{fig:mnist}.

\subsection{Settings}
In order to evaluate how the radial transform can increase diversity of a dataset and help the deep learning model to improve the generalization performance, the standard AlexNet and GoogLeNet models are trained with three different datasets: D1) Original images with 20 and 40 samples per class; D2) Affine transformed $\psi(\cdot)$ images of the original images, where 100 images are made from each original image, totally two datasets of size 2,000 and 4,000 images per class. The transformation parameters are selected randomly; D3) Radial transformed $\phi(\cdot)$ images of the original images, where 100 images are made from each original image, totaling two datasets of size 2,000 and 4,000 images per class. The pole is selected randomly in the original image. The datasets are made for MNIST and medical multimodal images, totaling six datasets.

The AlexNet and GoogLeNet models are trained using Stochastic Gradient descent over 50 iterations with exponential decay learning rate initialized to 0.01 and 0.001 for models trained with original and radial or Affine transformed images,  respectively. The parameters are selected based on grid search. The training dataset is shuffled to avoid sensitivity of the models to training order. The models are cross-validated over 30 independent experiments and a statistical test is conducted on the results. The validation and test datasets have 1,000 images each, with identical number of samples per class.

 \begin{table*}[!htp]
\renewcommand{\tabcolsep}{1pt}
\captionsetup{font=small}
%\scriptsize
\caption{The accuracy (``$\upsilon$" in $\%$), top-one probability confidence value (``$\kappa$" in $\%$), and converged-in iteration ($\xi$) of AlexNet and GoogLeNet models on the test dataset trained with original, Affine transform, and radial transform augmented MNIST and medical multimodal images. ``Std" is the standard deviation. The best result is in boldface.}
\vspace{-7mm}
\begin{center}
\begin{adjustbox}{width=0.9\textwidth}
\begin{tabular}{|c|c|c|c|c|c|c|c|c|c|c|c|c|c|}
\hline
\multirow{4}{*}{Model}       & \multirow{4}{*}{Transform} & \multicolumn{12}{c|}{Number of Original Images per Class}                                                                                       \\ \cline{3-14} 
                                  &                        & \multicolumn{6}{c|}{MNIST}                                & \multicolumn{6}{c|}{\begin{tabular}[c]{@{}c@{}}Medical Multimodal\end{tabular}} \\ \cline{3-14} 
                                  &                        & \multicolumn{3}{c|}{20}         & \multicolumn{3}{c|}{40} & \multicolumn{3}{c|}{20}                               & \multicolumn{3}{c|}{40}    \\ \cline{3-14} 
                                  &                        & $\upsilon$              & $\kappa$       & $\xi$       & $\upsilon$          & $\kappa$       &$\xi$   & $\upsilon$                         & $\kappa$                    &$\xi$     & $\upsilon$            & $\kappa$      &$\xi$     \\ \hline
              
                    \multirow{3}{*}{GoogLeNet}     &   Original & 45.62$\pm$1.91 & 28.67$\pm$2.28 &        48    &69.39$\pm$0.73  &67.00$\pm$0.69 &42 &83.00$\pm$2.51            & 78.93$\pm$2.32            &       40       &    98.33$\pm$0.81      & 95.95$\pm$0.81 & 26 \\ \cline{2-14}

                                  & Affine              &        11.16$\pm$0.10        &      12.86$\pm$0.01         &    2        &      59.11$\pm$0.56      &    56.88$\pm$0.61  &        19                   &       48.28$\pm$0.72       &   55.21$\pm$0.70  &12&61.42$\pm$1.02&58.92$\pm$0.95&16       \\ \cline{2-14} 
                                  
                                  &       Radial        & 97.98$\pm$1.39 & 99.39$\pm$1.22 &    13        &   91.96$\pm$0.89  &92.72$\pm$0.93 &  15     & 94.48$\pm$1.36            & 97.33$\pm$1.42            &     8         &    99.21$\pm$0.92 &99.14$\pm$0.98&8         \\ \hline
                                                     
\multirow{3}{*}{AlexNet}         &   Original              & 82.26$\pm$1.06 & 84.35$\pm$1.28 &   33    &  83.60$\pm$0.51   &    83.30$\pm$0.56 & 40& 89.01$\pm$1.28            & 88.14$\pm$1.39       &  21   &   98.04$\pm$ 0.61          &  98.33$\pm$     0.58 &   25  \\ \cline{2-14} 

        &   Affine              &       32.52$\pm$0.43         &     33.54$\pm$0.48           &    43        &           38.50$\pm$0.52 &        37.17$\pm$0.59    &  48    &      52.80$\pm$0.78        &   46.61$\pm$0.89&15 & 47.66$\pm$0.82&43.46$\pm$0.89&  7        \\ \cline{2-14} 

   &     Radial            & \textbf{98.29$\pm$0.96} & \textbf{98.57$\pm$0.98 }&    \textbf{19}        &    \textbf{95.18$\pm$0.74}    &\textbf{94.36$\pm$0.69} & \textbf{17}   & \textbf{97.05$\pm$1.01} & \textbf{99.34$\pm$1.21}  &  \textbf{ 4}    &     \textbf{99.54$\pm$0.66}  &\textbf{98.01$\pm$0.51}&  \textbf{4}     \\ \hline
                                                         
\end{tabular}
\end{adjustbox}
\label{T:lossAll}
\end{center}
        \vspace{-4mm}
\end{table*}

  \begin{table*}[t]
\renewcommand{\tabcolsep}{1pt}
\captionsetup{font=small}
%\scriptsize
\caption{The accuracy (``$\upsilon$" in $\%$) and top-one probability confidence value (``$\kappa$" in $\%$) of AlexNet and GoogLeNet models on the test dataset trained with original, Affine transform, and radial transform augmented medical multimodal images. The dataset size is 20 and 2,000 per class for the original and transformed images, respectively.}
\vspace{-7mm}
\begin{center}
\begin{adjustbox}{width=0.92\textwidth}

\begin{tabular}{|c|c|c|c|c|c|c|c|c|c|c|c|c|c|c|c|c|c|c|c|c|c|}
\hline
\multirow{3}{*}{Model} &\multirow{3}{*}{Transform} &\multicolumn{18}{c|}{Category} \\ \cline{3-20}

                                   & 									&\multicolumn{2}{c|}{Coronal Abd.}  &\multicolumn{2}{c|}{Trans. Abd.}&\multicolumn{2}{c|}{Sagittal Abd.} &\multicolumn{2}{c|}{Breast MRI} &\multicolumn{2}{c|}{Chest X-Ray} &\multicolumn{2}{c|}{Head CT} &\multicolumn{2}{c|}{Lung CT} &\multicolumn{2}{c|}{Mammogram} &\multicolumn{2}{c|}{Pelvis MRI}   \\ \cline{3-20}

                                   & 									&$\upsilon$ & $\kappa$&
                                   $\upsilon$ & $\kappa$&$\upsilon$ & $\kappa$&$\upsilon$ & $\kappa$&$\upsilon$ & $\kappa$&$\upsilon$ & $\kappa$&$\upsilon$ & $\kappa$&$\upsilon$ & $\kappa$&$\upsilon$ & $\kappa$ \\ \hline
                                   
\multirow{3}{*}{GoogLeNet}&Original&92.18	&	78.21	&	89.99	&	89.98	&	30.78	&	53.12	&	94.44	&	67.13	&	75.24	&	89.13	&	75.24	&	89.13	&	96.24	&	72.78	&	98.03	&	88.19	&	94.87	&	82.70		\\ \cline{2-20}		
			
													& Affine & 1.00&	13.73&1.00&16.83&48.00&52.26&95.82&83.94&79.22&70.29&82.53&78.49&27.00&52.34&65.00&74.83&35.00&48.83\\ \cline{2-20}
													
  & Radial & 98.12&	99.59	&	\textbf{99.90}	&	94.92	&	82.53	&	86.18	&	97.25	&	99.15	&	85.17	&	97.84	&\textbf{	90.17}	&	99.37	&	99.33	&	99.14	&	\textbf{99.79}	&	99.82	&	98.06	&	99.99\\ \hline

\multirow{3}{*}{AlexNet}& Original& 96.00	&	91.51	&	95.10	&	97.00	&	54.28	&	89.25	&	97.00	&	99.97	&	93.18	&	92.69	&	87.91	&	89.07	&	92.37	&	91.80	&	94.03	&	90.04	&	91.30	&	90.87		\\ \cline{2-20}

										& Affine & 1.00	&	12.67	&	0.00&0.00& 69.00	&	65.22& 96.03	&84.20&	71.45	&69.19&	 81.77 &85.28&54.00&66.29&68.00&73.92&34.00&52.91	\\ \cline{2-20}
										
										& Radial & \textbf{98.96}	&	\textbf{99.60}	&	99.70	&	\textbf{99.91}	&	\textbf{86.31}	&	\textbf{95.38}	&	\textbf{99.90}	&	\textbf{99.98}	&	\textbf{99.99}	&	\textbf{99.99}	&	89.17	&	\textbf{99.38}	&\textbf{	100.00	}&\textbf{	99.99}	&	99.49	&	\textbf{99.91}	&	\textbf{100.00}	&	\textbf{99.99}	\\  \hline

\end{tabular}
\end{adjustbox}
\label{T:lossmultimodal}
\end{center}
\vspace{-4mm}
\end{table*}
  \begin{table*}[!htp]
\renewcommand{\tabcolsep}{1pt}
\captionsetup{font=small}
%\scriptsize
\caption{The accuracy (``$\upsilon$" in $\%$) and top-one probability confidence value (``$\kappa$" in $\%$) of AlexNet and GoogLeNet models on the test dataset trained with original, Affine transform, and radial transform augmented MNIST images. The dataset size is 20 and 2,000 per class for the original and transformed images, respectively.}
\vspace{-7mm}
\begin{center}
\begin{adjustbox}{width=0.94\textwidth}

\begin{tabular}{|c|c|c|c|c|c|c|c|c|c|c|c|c|c|c|c|c|c|c|c|c|c|c|c|}
\hline
\multirow{3}{*}{Model} &\multirow{3}{*}{Transform} &\multicolumn{20}{c|}{Category} \\ \cline{3-22}

                                   & 									&\multicolumn{2}{c|}{0}  &\multicolumn{2}{c|}{1}&\multicolumn{2}{c|}{2} &\multicolumn{2}{c|}{3} &\multicolumn{2}{c|}{4} &\multicolumn{2}{c|}{5} &\multicolumn{2}{c|}{6} &\multicolumn{2}{c|}{7} &\multicolumn{2}{c|}{8} & \multicolumn{2}{c|}{9}    \\ \cline{3-22}

                                   & 									&$\upsilon$ & $\kappa$&$\upsilon$ & $\kappa$&$\upsilon$ & $\kappa$&$\upsilon$ & $\kappa$&$\upsilon$ & $\kappa$&$\upsilon$ & $\kappa$&$\upsilon$ & $\kappa$&$\upsilon$ & $\kappa$&$\upsilon$ & $\kappa$ &$\upsilon$ & $\kappa$   \\ \hline
\multirow{3}{*}{GoogLeNet}&Original&21.29	&	22.60	&	96.14	&	89.98	&	16.69	&	19.57	&	48.44	&	18.84	&	66.22	&	33.27	&	23.49	&	17.22	&	48.08	&	17.59	&	72.93	&	30.17	&	25.96	&	16.96	&	37.04	&	20.57	\\ \cline{2-22}

& Affine & 0.54&13.19&90.00&	13.86&1.00&17.30&2.47&16.83&1.25&17.92&3.12.00&15.24&3.01
&15.92&0.00&0.00&1.20&17.91&1.00&19.30 \\ \cline{2-22}	
			
  & Radial & 98.25	&	\textbf{99.35}	&	97.24	&	98.98	&	98.22	&	\textbf{99.41}	&	99.09	&	\textbf{99.74}	&	98.40	&	\textbf{99.61}	&	96.92	&	\textbf{99.37}	&	\textbf{98.98}	&	\textbf{99.59}	&	97.60	&	\textbf{99.25}	&	\textbf{97.64}	&	\textbf{99.46}	&	\textbf{97.54}	&	\textbf{99.23}	\\ \hline
  
\multirow{3}{*}{AlexNet}& Original	&	83.65	&	90.01	&	96.24	&	95.77	&	79.62	&	85.57	&	60.19	&	71.31	&	92.67	&	89.71	&	59.99	&	74.45	&	89.55	&	84.67	&	90.73	&	91.74	&	82.65	&	77.36	&	87.30	&	82.87	\\ \cline{2-22}

& Affine & 57.75&  46.81 & 87.46 &40.15 & 9.78  &46.00 &21.79  &57.47 &26.55   & 30.03&12.32  &47.24& 17.16 & 35.06&26.65 & 41.25&33.11 &37.62& 32.70&45.24 \\ \cline{2-22}

										& Radial & \textbf{99.37	}&	99.02	&	\textbf{98.76}	&	\textbf{99.21}	&	\textbf{98.45}	&	98.38	&	\textbf{99.52}	&	99.26	&	\textbf{99.22}&	98.74	&	\textbf{97.34}	&	98.37	&	98.29	&	98.98	&	\textbf{97.90}	&	98.00	&	96.62	&	97.93	&	97.42	&	97.81\\  \hline

\end{tabular}
\end{adjustbox}
\label{T:lossmnist}
\end{center}
        \vspace{-5mm}
\end{table*}
% Please add the following required packages to your document preamble:
% \usepackage{multirow}

\subsection{Results Analysis}
Figure~\ref{fig:mnist} shows how the Affine transform and the proposed radial transform can augment an image to generate a new representation. The augmented images using Affine transform show this transform can preserves points, maps a line to a line, and preserves parallel lines such as the head and tail of the number 5's image, ribs in the chest X-ray, and left and right obturator internus of pelvis MRI. This transform also preserves ratios of distances between points lying on a straight line. For example, the distance between the two sides (branches) of the number 4's image at top and bottom, fatty tissues in the mammogram, and Cerebellar hemisphere in the head computed tomography (CT). However, this transform may result in loss of resolution or a part of image as shown for the head CT, and chest X-ray. The resulted images from radial transform show that it preserves the local and global spatial features in the neighborhood of the pole. This transform up-samples the pixels sitting in close spatial proximity of the pole and down-samples the pixels distant to the spatial neighborhood of the pole. Despite the Affine transform, the radial transform does not necessarily preserve the parallel lines or distances between two specific point. However, it defines a logical relationship among the pixels based on the sampling in polar coordinate system. 

The accuracy ($\upsilon$), top-one probability confidence ($\kappa$), and converged-in iteration ($\xi$) of AlexNet and GoogLeNet are presented in Table~\ref{T:lossAll}. The performance values per image class for original dataset size of 20 and corresponding Affine and radial transformed images are presented in Tables ~\ref{T:lossmultimodal}~and~\ref{T:lossmnist} (due to lack of space, the data size of 20 is only presented). The accuracy of class $c$ is defined as $\upsilon_{c}=\sum_{s=1}^{|S|}\mathbb{I} [c=arg\: max(p_{s,1},...,p_{s,C})]/|S|$ where $S$ is the test dataset, $C$ is the number of classes, $p_{c,s}$ is the classification probability of the data sample $s$ for class $c$, and $\mathbb{I}[x]$ is defined to be 1 if $x$ is true, and 0 if it is false. The top-one probability confidence of class $c$ is defined as $\kappa_{c}=(\sum_{s=1}^{|S|}p_{s,c})/|S|.$

\begin{figure}[!t]
\captionsetup{font=small}
\centering
\vspace{-5mm}
\includegraphics[width=0.45\textwidth]{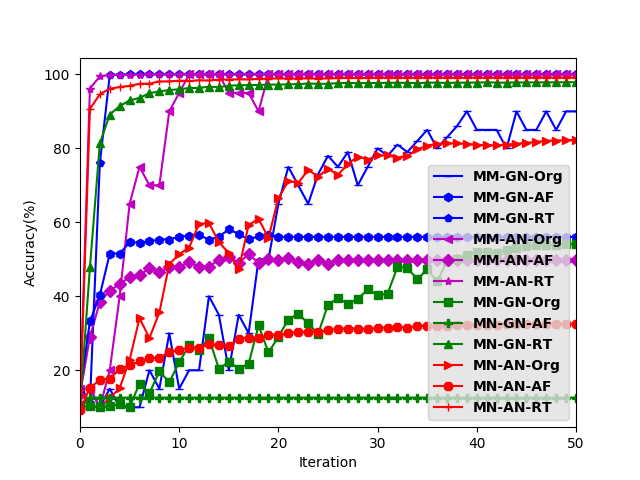}
 \vspace{-3mm}       
\caption{Accuracy of the AlexNet (AN) and GoogLeNet (GN), trained with original (Org), Affine transformed (AF), and radial transformed (RT) images on MNIST (MN) and medical multimodal (MM) validation datasets. }
\label{fig:acc_loss}
\vspace{-6mm}
\end{figure}

%$100\times |\{c: arg\: max(p_{s,1},...,p_{s,C})=c \:\: \forall \: s\in S\}|/ |S|$

The results clearly show that the models trained with radial transformed data have better performance. At competitive accuracy, the confidence of models trained with radial transform is greater. The difference in accuracy of the trained models is more obvious for the MNIST dataset, likely due to the correlation among the medical images such as between Transverse abdomen MRI and Sagittal abdomen MRI.

The accuracy of the model on the validation dataset through training iterations for a single experiment is presented in Figure~\ref{fig:acc_loss}. The converged-in iteration $\xi$ in Table~\ref{T:lossAll} and this figure show that the models trained with radial transformed images D3 converge faster with higher accuracy than models trained with D1 and D2. AlexNet and GoogLeNet trained with very limited original images show fluctuation of validation accuracy during training. GoogLeNet has more fluctuation, particularly due to having more number of free parameters than AlexNet. The same models trained with augmented images using radial transform show smoother convergence and less fluctuation of the validation accuracy.

\section{Conclusion}
\label{sec:conclusion}
Successful training of deep neural networks requires a large quantity of balanced data. In practice, most of the datasets are imbalanced and often very limited data is available for certain classes in a dataset. In this paper, we propose image augmentation using radial transform in the polar coordinate system to facilitate training of deep neural networks. This method preserves the information content of the original image, but improves the diversity of the training dataset, resulting in improved generalization performance of the neural network.

%Our experiments show that these models are highly sensitive to %training order and parameters initialization when very little data is %available. 

%\vfill\pagebreak

\bibliographystyle{IEEEbib}
\bibliography{strings,mybibfile}

\end{document}